\documentclass[12pt]{article}

\usepackage{times}
\usepackage{hyperref}
\usepackage{graphicx}
\usepackage{amsmath}
\usepackage{amsfonts}
\usepackage{float}

\newenvironment{sciabstract}{%
\begin{quote} \bf}
{\end{quote}}

\newcounter{lastnote}

\title{Using human brain activity to guide machine learning}

\author
{Ruth C. Fong,$^{1,3,\dag}$ Walter J. Scheirer,$^{2,3\dag}$ David D. Cox$^{3\ast}$\\
\\
\normalsize{$^{1}$Department of Engineering Science, University of Oxford}\\
\normalsize{Information Engineering Building, Oxford OX1 3PJ, United Kingdom}\\
\\
\normalsize{$^{2}$Department of Computer Science and Engineering, University of Notre Dame}\\
\normalsize{Fitzpatrick Hall of Engineering, Notre Dame, IN, 46556, USA}\\
\\
\normalsize{$^{3}$Department of Molecular and Cellular Biology, School of Engineering and Applied Sciences}\\
\normalsize{and Center for Brain Science, Harvard University.}\\
\normalsize{52 Oxford St., Cambridge, MA, 02138, USA}\\
\\
\normalsize{$^\ast$To whom correspondence should be addressed; E-mail: davidcox@fas.harvard.edu.}\\
\normalsize{$^\dag$R.C. Fong and W.J. Scheirer contributed equally to this work.}
}

\date{}

\begin{document} 


\baselineskip24pt


\maketitle

\pagebreak

\section*{Abstract}


\begin{sciabstract} 
Machine learning is a field of computer science that builds algorithms that learn. In many cases, machine learning algorithms are used to recreate a human ability like adding a caption to a photo, driving a car, or playing a game.
While the human brain has long served as a source of \emph{inspiration} for machine learning, little effort has been made to directly use data collected from working brains as a \emph{guide} for machine learning algorithms. 
Here we demonstrate a new paradigm of ``neurally-weighted'' machine learning, which takes fMRI measurements of human brain activity from subjects viewing images, and infuses these data into the training process of an object recognition learning algorithm to make it more consistent with the human brain.
After training, these neurally-weighted classifiers are able to classify images without requiring any additional neural data.
We show that our neural-weighting approach can lead to large performance gains when used with traditional machine vision features, as well as to significant improvements with already high-performing convolutional neural network features.
The effectiveness of this approach points to a path forward for a new class of hybrid machine learning algorithms which take both inspiration and direct constraints from neuronal data.
\end{sciabstract}

\pagebreak

\section*{Introduction}

Recent years have seen a renaissance in machine learning and machine vision, led by neural network algorithms that now achieve impressive performance on a variety of challenging object recognition and image understanding tasks~\cite{russakovsky2014imagenet, karpathy2014large, karpathy2015deep}.
Despite this rapid progress, the performance of machine vision algorithms continues to trail humans in many key domains, and tasks that require operating with limited training data or in highly cluttered scenes are particularly difficult for current algorithms~\cite{borji_2012, borji_2014, bendale2015towards,hoiem2012diagnosing}. 
Moreover, the patterns of errors made by today's algorithms differ dramatically from those of humans performing the same tasks~\cite{Scheirer_2014_TPAMIa, pramod2016computational}, and current algorithms can be ``fooled'' by subtly altering images in ways that are imperceptible to humans, but which lead to arbitrary misclassifications of objects~\cite{Szegedy_2013, Szegedy_2015, Nguyen_2015_CVPR}.
Thus, even when algorithms do well on a particular task, they do so in a way that differs from how humans do it and that is arguably more brittle.

The human brain is a natural frame of reference for machine learning, because it has evolved to operate with extraordinary efficiency and accuracy in complicated and ambiguous environments. 
Indeed, today's best algorithms for learning structure in data are artificial neural networks~\cite{krizhevsky2012imagenet, DBLP:journals/corr/JiaSDKLGGD14, Mnih_2015}, and strategies for decision making that incorporate cognitive models of Bayesian reasoning~\cite{Tenenbaum_2006} and exemplar learning~\cite{Jakel_2009} are prevalent. 
There is also growing overlap between machine learning and the fields of neuroscience and psychology: In one direction, learning algorithms are used for fMRI decoding~\cite{kay2008identifying, mitchell2008predicting, naselaris2009bayesian, reddy2010reading}, neural response prediction~\cite{yamins2013hierarchical, yamins2014performance, kriegeskorte2008representational,arxiv14_Agrawal,cichy2016deep}, and hierarchical modeling~\cite{Riesenhuber_1999, serre2007feedforward, pinto2009high}. Concurrently, machine learning algorithms are also leveraging biological concepts like working memory~\cite{graves2016hybrid}, experience replay~\cite{mnih2015human}, and attention~\cite{gregor2015draw,xu2015show} and are being encouraged to borrow more insights from the inner workings of the human brain~\cite{lake2016building}.
Here we propose an even more direct connection between these fields: we ask if we can improve machine learning algorithms by explicitly guiding their training with measurements of brain activity, with the goal of making the algorithms more human-like.



\begin{figure}[!ht]
\centering
\includegraphics[width=1\linewidth]{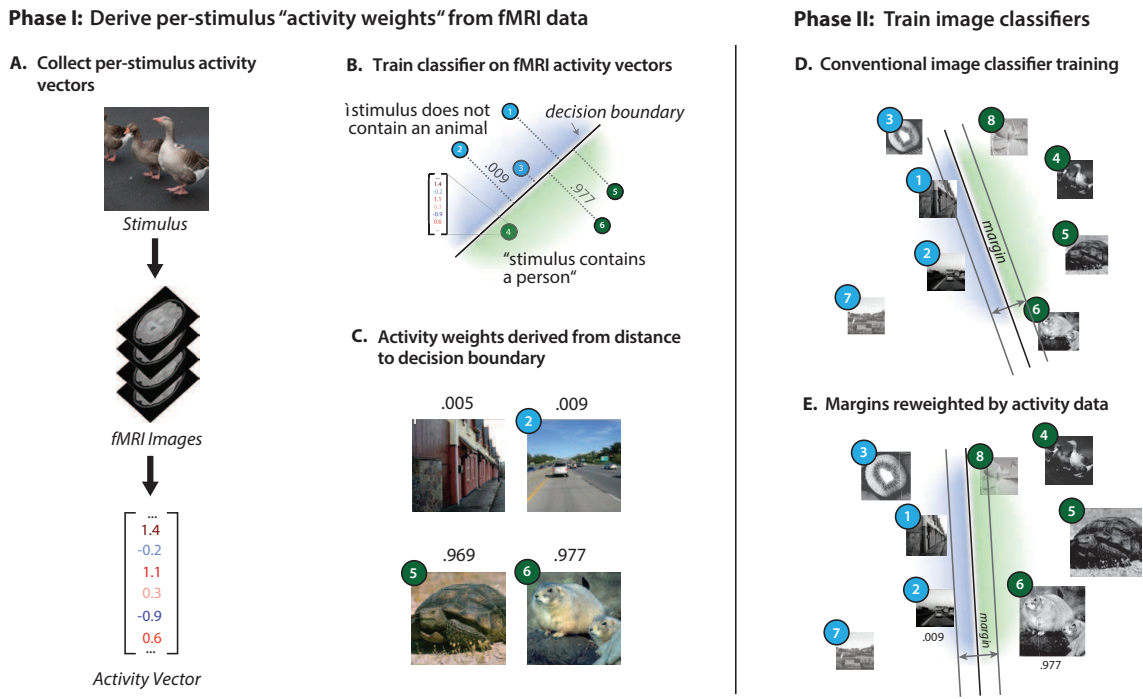}
\caption{Experimental workflow for biologically-informed machine learning using fMRI data. 
\textbf{(A)} fMRI was used to record BOLD voxel responses of one subject viewing $1,386$ color images of natural scenes, providing labelled voxels in several conventional, functional ROIs (i.e., EBA, FFA, LO, OFA, PPA, RSC, and TOS)~\cite{stansbury2013natural}. 
\textbf{(B)} For a given binary classification task category (e.g., whether a stimulus contains a person), the visual stimuli and voxel activity data were split into training and test sets. 
An SVM classifier was trained and tested on voxel activity alone. 
\textbf{(C)} To generate activity weights, classification scores, which roughly correspond to the distance of a sample from the decision boundary in \textbf{(B)}, were transformed into a probability value via a logistic function~\cite{platt_2000}. 
\textbf{(D, E)} The effects of using activity weights were assessed by training and testing two classification models on image features of the visual stimuli: 
\textbf{(D)} One SVM classifier used a loss function (e.g., hinge loss [HL]) that equally weights the misclassification of all samples as a function of distance from the SVM's own decision boundary. 
\textbf{(E)} Another SVM classifier used a modified loss function (e.g., activity weighted loss [AWL]) that penalizes more aggressively the misclassification of samples with large activity weights. 
In training, these classifiers in \textbf{(D)} and \textbf{(E)} only had access to activity weights generated in \textbf{(C)}; in testing, the classifiers used no neural data and made predictions based on image features alone.}
\label{fig:fig1}
\end{figure}

Our strategy is to bias the solution of a machine learning algorithm so that it more closely matches the internal representations found in visual cortex. 
Previous studies have constrained learned models via human behavior~\cite{Scheirer_2014_TPAMIa, chen_2010}, and one work introduced a method to determine a mapping from images to ``brain-like'' features extracted from EEG recordings~\cite{spampinato2016deep}. Furthermore, recent advances in machine learning have focused on improving feature representation, often in a biologically-consistent way~\cite{cichy2016deep}, of different kinds of data. However, no study to date has taken advantage of measurements of brain activity to guide the decision making process of machine learning. While our understanding of human cognition and decision making is still  limited, we describe a method with which we can leverage the human brain's robust representations to guide a machine learning algorithm's decision boundary. 
Our approach weights how much an algorithm learns from each training exemplar, roughly based on the ``ease'' with which the human brain appears to recognize the example as a member of a class (i.e., an image in a given object category). This work builds on previous machine learning approaches that weight training~\cite{bengio2009curriculum,Scheirer_2014_TPAMIa}, but here we propose to do such weighting using a separate stream of data, derived from human brain activity. 


Below, we describe our biologically-informed machine learning paradigm in detail, outline an implementation of the technique, and present results that demonstrate its potential to learn more accurate, biologically-consistent decision-making boundaries. 
We trained supervised classification models in visual object categorization, weighting individual training images by values derived from fMRI recordings in human visual cortex viewing those same images; once trained, these models classify images without the benefit of neural data.
These models were then evaluated for improvement in baseline performance as well as analyzed to understand which regions of interest (ROIs) in the brain had greater impact on performance.

\section*{Results}

Visual cortical fMRI data were taken from a previous study conducted by the Gallant lab at Berkeley~\cite{stansbury2013natural}. 
One adult subject viewed $1,386$ color $500\times500$ pixel images of natural scenes, while being scanned in a $3.0$ Tesla ($3$T) magnetic resonance imaging (MRI) machine. 
After fMRI data preprocessing,  response amplitude values for $67,600$ voxels were available for each image. 
From this set of voxels, $3,569$ were labeled as being part of one of thirteen visual ROIs, including those in the early visual cortex. Seven of these regions associated with higher-level visual processing were chosen for use in object classification tasks probing the semantic understanding of visual information: extrastriate body area (EBA), fusiform face area (FFA), lateral occipital cortex (LO), occiptal face area (OFA), parahippocampal place area (PPA), retrosplenial cortex (RSC), transverse occipital sulcus (TOS). $1,427$ voxels belonged to these regions.

In machine learning, loss functions are used to assign penalties for misclassifying data; then, the objective of the algorithm is to minimize loss. Typically, a \textit{hinge loss} function (Equation~\ref{eqn:hinge_loss}) is used for classic maximum-margin binary classifiers like Support Vector Machine (SVM) models~\cite{cortes1995support}:
\begin{equation}
\label{eqn:hinge_loss}
\phi_h(z) = \max(0, 1-z)
\end{equation}
where $z = y \cdot f(x)$, $y \in \mathbb{N}$ is the true label, and $f(x) \in \mathbb{R}$ is the predicted output; thus, $z$ denotes the correctness of a prediction. The HL function assigns a penalty to all misclassified data that is proportional to how erroneous the prediction is. 

However, incorporating brain data into a machine learning model relies on the assumption that the intensity of a pattern of activation in a region represents the neuronal response to a visual stimulus. A strong response signals that a stimulus is easy for a subject to recognize, while a weaker response indicates that the stimulus is more difficult to recognize~\cite{gauthier2000expertise}. Here, the proposed \textit{activity weighted loss} (AWL) function (Equation~\ref{eqn:neuro_weighted_loss}) embodies this strategy by proportionally penalizing misclassified training samples based on any inconsistency with the evidence of human decision making found in the fMRI measurements, in addition to using the typical HL penalty:
\begin{equation}
    \label{eqn:neuro_weighted_loss}
    \phi_\psi(x,z) = \max(0,(1-z) \cdot M(x,z))
\end{equation}
where
\begin{equation}
\label{eqn:awl_mapping}
    M(x,z) = \begin{cases} 1 + c_x, & \text{if } z < 1\\1, & \text{otherwise}\end{cases}
\end{equation}
and $c_x \geq 0$ is an activity weight derived from fMRI data corresponding to $x$.

A large activity weight $c_x$ corresponds to a strong neuronal response pattern to visual stimulus $x$; thus, AWL penalizes more aggressively the misclassification of stimuli that humans recognize consistently with ease. With this formulation, not all training samples require an fMRI-generated activity weight. Note that $c_x = 0$ reduces the AWL function to a HL function and can be assigned to samples for which fMRI data is unavailable. AWL is inspired by previous work~\cite{Scheirer_2014_TPAMIa}, which introduced a loss function that additively scaled misclassification penalty by information derived from behavioral data. AWL replaces the standard HL function (Equation~\ref{eqn:hinge_loss}) in the objective of the SVM algorithm, which does not have access to any information other than a feature vector and an arbitrary class label for each training sample in its original form. 

Experiments were conducted for the $127$ ways that the seven higher-level visual cortical regions could be combined. 
In each experiment, for a given combination of ROIs and a given object category, the following procedure was carried out (Figure~\ref{fig:fig1}): 
1. Generate activity weights by calibrating the scores of a Radial Basis Function (RBF) kernel SVM classifier, trained on the training voxel data for the combination, into probabilities via a logistic transformation (Figures~\ref{fig:fig1}A-\ref{fig:fig1}C) ~\cite{platt_2000}. 
2. Create five balanced classification problems (Figure S1). For each balanced problem and a set of image descriptors, train and test two SVM classification models with an RBF-kernel --- one that used the HL function and another that used an AWL function conditioned on the activity weights from the first step (Figures~\ref{fig:fig1}D and~\ref{fig:fig1}E). 
Two image features were considered: the Histogram of Oriented Gradients (HOG) is a handcrafted feature that is approximately V1-like~\cite{dalal_2005,vedaldi08vlfeat}; convolutional neural networks (CNN) are learned feature representations that approximate several additional layers of the ventral stream~\cite{yamins2013hierarchical,DBLP:journals/corr/JiaSDKLGGD14}. 
Experiments were performed for four object categories: humans, animals, buildings, and foods.

\begin{figure}[!ht]
\centering
\includegraphics[width=0.95\linewidth]{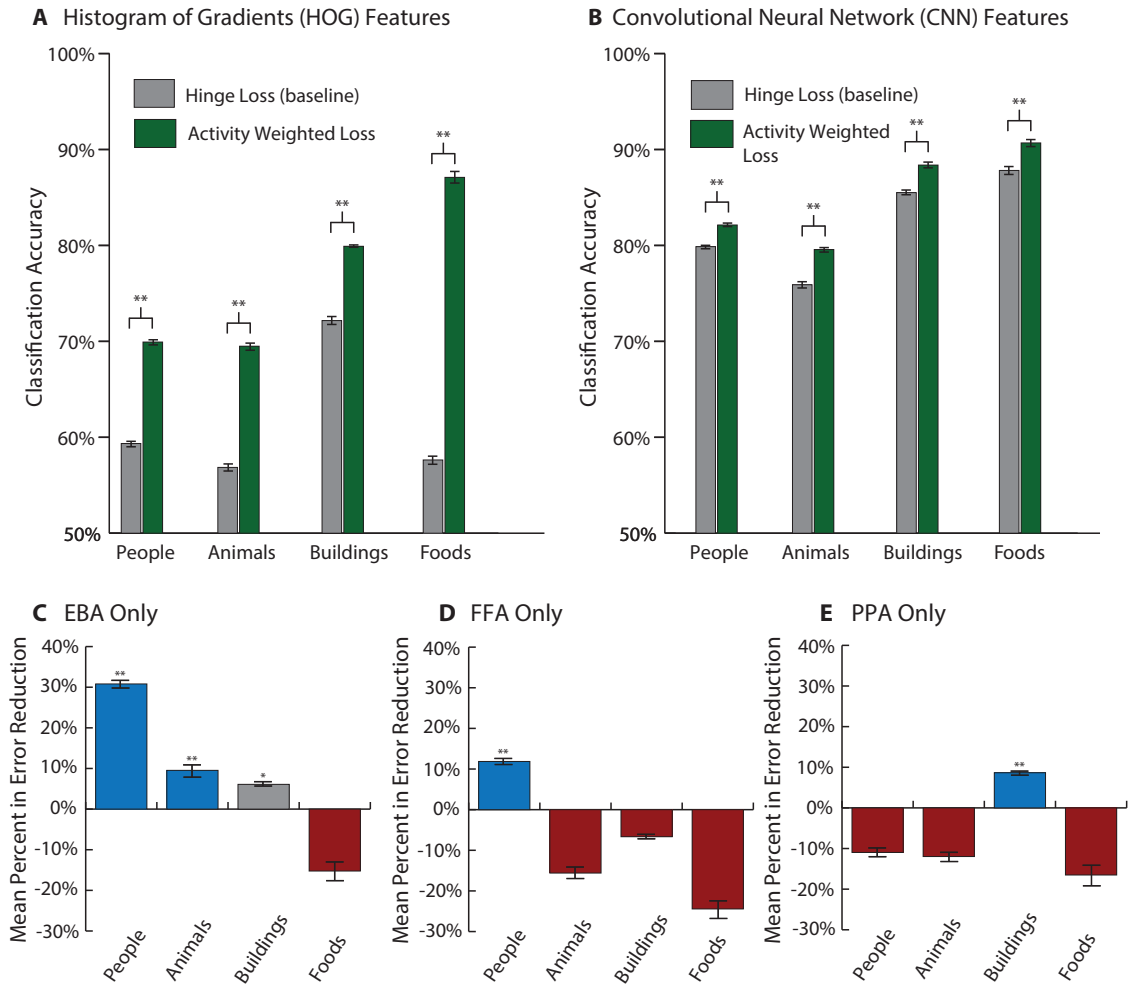}
\caption{Results showing the effect of conditioning classification models for four visual classes on fMRI data during supervised training (see Figure~\ref{fig:fig1}). \textbf{(A,B)} Side-by-side comparisons of the mean classification accuracy between models that were trained using either \textbf{(A)} HOG features or \textbf{(B)} CNN features and either a hinge loss (HL) or activity weighted loss (AWL) function. These graphs show results of experiments that generated activity weights from seven higher-level visual ROIs (i.e., EBA, FFA, LO, OFA, PPA, RSC, TOS). In all  cases, models trained using AWL were significantly better ($p < 0.01$ via paired, one-tailed t-testing). While using AWL loss reduces misclassification error using both features, it particularly improves the performance of handcrafted HOG features. \textbf{(C-E)} Mean error reductions gained by switching from HL to AWL loss when using conditioning classifies on brain activity from individual ROIs (i.e., EBA, FFA, or PPA) show that certain areas produce significantly better results for the specific categories they are selective for. Error bars are standard error over 20 trials in all cases.}
\label{fig:fig2}
\end{figure}

We demonstrate that using activity weights derived from all of the higher visual cortical regions significantly improves classification accuracy across all four object categories via paired, one-tailed testing (Figures~\ref{fig:fig2}A and ~\ref{fig:fig2}B). 
A substantial amount of fMRI decoding literature focuses on three ROIs: EBA, FFA, and PPA~\cite{cox_2003,kriegeskorte_2007,Spiridon_2006}. 
This is in part because these three regions are thought to respond to visual cues of interest for the study of object recognition: body parts, faces, and places respectively. 
Given the overlap between these visual cues and the four object categories used, we hypothesized that activity weights derived from brain activity in these three regions would significantly improve classification accuracy for the humans, animals, and buildings categories only in instances where a response would be expected. 
For example, PPA was expected to improve the buildings category but to have little, if any, effect on the humans category (Figure~\ref{fig:fig2}E). 
A comparison of models that used activity weights based on brain activity from these three regions and models that used no activity weights aligns with the neuroimaging literature (Figures~\ref{fig:fig2}C-~\ref{fig:fig2}E). 
Classification accuracy significantly improved not only when activity weights were derived from voxels in all seven ROIs or from voxels in the individual EBA, FFA, and PPA regions but also when activity weights were derived from voxels in most of the $127$ ROI combinations (see Figure S2). 
We observed that adding fMRI-derived activity weights provided a large improvement to models using HOG features compared to those using CNN features (Figures~\ref{fig:fig2}A and~\ref{fig:fig2}B). 
These results suggest that improvements in decision making (e.g., the use of salient activity weights based on brain activity) may be able to compensate for poor feature representation (e.g., HOG features). They also imply that some of the information carried by activity weights may already be latently captured in CNN features. Despite their relatively smaller performance gains, activity weighted classifiers for CNN features still demonstrate that the state-of-the-art representation, which is often praised as being inspired by the mammalian ventral steam, does not fully capture all the salient information embedded in internal representations of objects in the human brain.

\begin{figure}[!t]
\centering
\includegraphics[width=1\linewidth]{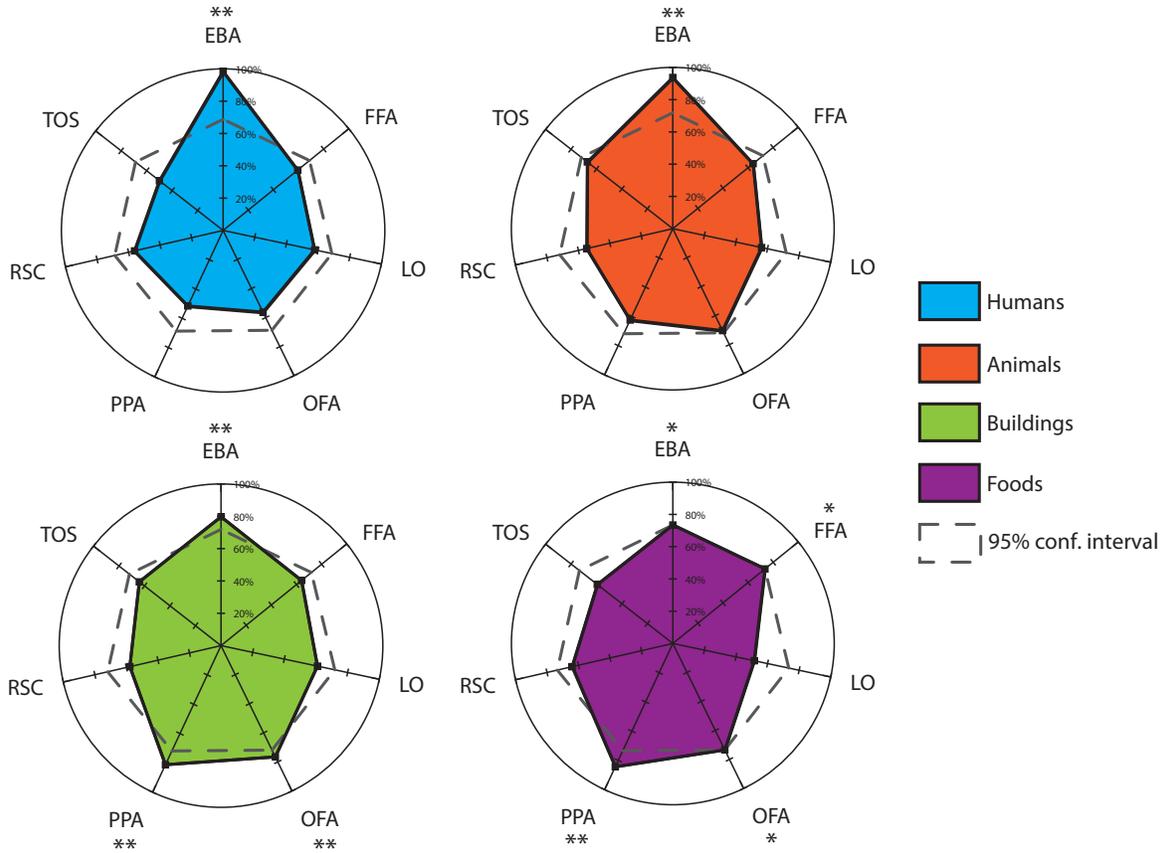}
\caption{Statistical influence of each ROI in object classification models using AWL and HOG features. 
In each graph, the fraction of the 64 ROI combinations containing a specific ROI that had a mean classification accuracy greater than that of all 127 sets of experiments is plotted. 
The threshold for the 95\% confidence interval ($p < 0.0004$) is overlaid, showing which ROIs significantly differed from the respective null distribution for each object category. 
Permutation tests and Bonferroni correction ($\alpha = 127$) were used.}
\label{fig:fig3}
\end{figure}

Additionally, statistical analysis by permutation was carried out to test whether the above-average accuracy rates observed in classification experiments for the humans and animals categories that included EBA, as well as in the experiments for the buildings and foods categories that included PPA, were statistically significant or products of random chance. For each object category and set of image features, a null distribution with $1,000,000$ samples was generated. Each sample in the null distribution reflects how often a random set of $64$ ROI combinations would have an above-average classification accuracy. The aim is to test the significance of individual ROIs in generating salient activity weights that yield above-average classification accuracy rates. Thus, these samples simulate randomly assigning ROI labels to the $127$ combinations. If individual ROIs did not significantly contribute to the above-average accuracy rates observed, above-average accuracy rates of combinations that include specific ROIs falling near the mean of the null distribution should be observed. To generate each of the $1,000,000$ samples, $64$ ROI combinations were randomly selected. Then, a count was taken of how many of those $64$ randomly selected combinations have a mean classification accuracy that is greater than that of all $127$ sets of experiments corresponding to the $127$ total ROI combinations. A sample is normalized by dividing this count by $64$. Figure~\ref{fig:fig3} and Figure S3 show which ROIs significantly differed from the respective null distributions for each object category. This analysis more rigorously confirms the significance of the EBA region in improving the classification accuracy of the humans and animals categories and of the PPA region in improving the accuracy of the buildings and foods categories. Most notably, the EBA region dramatically exceeds the significance thresholds of the null distributions for humans and animals.

\section*{Discussion}
Our results provide strong evidence that information measured directly from the human brain can help a machine learning algorithm make better, more human-like decisions.
As such, this work adds to a growing body to literature that suggests that it is possible to leverage additional ``side-stream'' data sources to improve machine learning algorithms~\cite{Scheirer_2014_TPAMIa,vondrick2014acquiring}.
However, while measures of human behavior have been used extensively to guide machine learning via active learning~\cite{cohn1996active,kapoor2007active,jain2009active,settles2010active}, structured domain knowledge~\cite{NIPS2011_4199}, and discriminative feature identification~\cite{deng2016leveraging}, this study suggests that one can harness measures of the internal representations employed by the brain to guide machine learning.
We argue that this approach opens a new wealth of opportunities for fine-grained interaction between machine learning and neuroscience.

While this work focused on visual object recognition and fMRI data, the framework described here need not be specific to any one sensory modality, neuroimaging technique or supervised machine learning algorithm.  Indeed, the approach can be applied generally to any sensory modality, and could even potentially be used to study multisensory integration, with appropriate data collection.
Similarly, while fMRI has the advantage of measuring patterns of activity over large regions of the brain, one could also imagine applying our paradigm to neural data collected using other imaging methods in animals, including techniques that allow single cell resolution over cortical populations, such as two-photon imaging \cite{ohki2014vivo}.
Such approaches may allow more fine-grained constraints to be placed on machine learning, albeit at the expense of allowing the integration of data from smaller fractions of the brain.


 
There are several limitations with this first instantiation of the paradigm. First, we derived a scalar activity weight from high-dimensional fMRI voxel activity. This simple method yielded impressive performance gains and corresponded well to the notion of ease of recognition; however, much more meaningful information captured by the human brain is inevitably being ignored. Future biologically-informed machine learning research should focus on the development and infusion of low-dimensional activity weights, which may not only preserve more useful data but also reveal other dimensions that are important for various tasks, but are not yet learned by machine learning algorithms or captured in traditional datasets.

Similarly, while we demonstrated our biologically-informed paradigm using support vector machines, there is also flexibility in the choice of the learning algorithm itself. Our method can be applied to any learning algorithm with a loss formulation as well as extended to other tasks in regression and Bayesian inference. An analysis of different algorithms and their baseline and activity weighted performance could elucidate which algorithms are relatively better at capturing salient information encoded in the internal representations of the human brain~\cite{kriegeskorte2008representational}. 

Our paradigm currently requires access to biological data during training time that corresponds to the input data for a given task.  For instance, in this work, we used fMRI recordings of human subjects viewing images to guide learning of object categories.
Extending this work to new problem domains will require specific data from those problem domains, and this will in turn require either increased sharing of raw neuroimaging data, or close collaboration between neuroscientists and machine learning researchers.
While this investigation used preexisting data to inform a decision boundary, one could imagine even more targeted collaborations between neuroscientists and machine learning researchers that tailor data collection to the needs of the machine learning algorithm.
We argue that approaches such as ours provide a framework of common ground for such collaborations, from which both fields stand to benefit.



\section*{Methods}
\subsection*{fMRI Data Acquisition} The fMRI data used for the machine learning experiments presented in this paper are a subset of the overall data from a published study on scene categorization~\cite{stansbury2013natural}. All fMRI data were collected using a 3T Siemens Tim Trio MR scanner. For the functional data collection for one subject, a gradient-echo planar imaging sequence, combined with a custom fat saturation RF pulse, was used. Twenty-five axial slices covered occipital, occipitoparietal, and occipitotemporal cortex. Each slice had a $234 \times 234 mm^2$ field of view, $2.60 mm$ slice thickness, and $0.39 mm$ slice gap (matrix size = $104 \times 104$; TR = $2,009.9 ms$; TE = $35 ms$; flip angle = $74^\circ$; voxel size = $2.25 \times 2.25 \times 2.99 mm^3$). The experimental protocol was approved by the UC Berkeley Committee for the Protection of Human Subjects. All methods were performed in accordance with the relevant guidelines and regulations.

\subsection*{Data Set} The data set consisted of $1,386$ $500 \times 500$ color images of natural scenes, which the subject viewed while his brain activity was being recorded. These images were used as both stimuli for the fMRI data collection, and as training data for the machine learning experiments. Within this collection, the training set consisted of $1,260$ images and the testing set of $126$ images. Per-pixel object labels in the form of object outlines and semantically meaningful tags were available for each image. A subset of these labels were mapped to one of five object categories: humans, animals, buildings, foods, and vehicles. For each image and for each of the five object categories, if at least $20\%$ of an image's original pixel labels were part of a given object category, that image was tentatively labeled as a positive sample for that category. We sampled $646$ images that were labelled with a single object category. There were $219$ humans images, $180$ animals images, $151$ buildings images, $59$ foods images, and $37$ vehicles images (a category, due to its small size, that only contributed negative examples).

\subsection*{fMRI Data Preprocessing} To perform motion correction, coregistration, and reslicing of functional images, the SPM8 package~\cite{SPM8} was used. Custom MATLAB~\cite{Matlab} software was used to perform all other preprocessing of functional data. To constrain the dimensionality of the fMRI data, the time series recordings for each voxel were reduced to a single response amplitude per image by deconvolving each time course from the stimulus design matrix. See Stansbury et al.~\cite{stansbury2013natural} for additional details about the fMRI data acquisition and preparation that are not directly related to the machine learning experiments we describe in this work.

\subsection*{fMRI Activity Weight Calculation} All of the fMRI data were scaled to bring the value of each dimension within the range of $[0,1]$ for RBF SVM training. For each voxel, we calculated the minimum and maximum response amplitude across all $1,260$ original training samples. All voxels for the $646$ images used in our experiments were then scaled using Equation~\ref{eq:scaling}, where $x_{ij}$ is the $j$-th sample's response amplitude for voxel $i$, $\vec{x_i}$ is a $646$-dimensional vector with the response amplitudes of all samples for voxel $i$, and $x_{ij}'$ is the $j$-th sample's rescaled amplitude for voxel $i$.
\begin{equation}
\label{eq:scaling}
x_{ij}' = \frac{x_{ij} - \min{(\vec{x_i})}}{\max{(\vec{x_i})} - \min{(\vec{x_i})}}
\end{equation}
The main challenge of generating weights from brain activity (i.e., activity weights) lies in reducing high-dimensional, nonlinear data to a salient, lower-dimensional signal of ``learnability.'' The supervised machine learning formulation used in this work requires a single real valued weight per training sample for a loss function (described below). Activity weights were computed by using a logistic transformation~\cite{platt_2000} to calibrate the scores from SVMs with RBF kernels trained on brain activity. For each object category and for all voxels from a given combination of ROIs, we made use of all the positive samples for that object category as well as all the samples that are negative for all object categories; together, these are the aforementioned $646$ samples (i.e., clear sample set). Only activity weights for this subset of a partition's training set, as opposed to annotations for all $1,386$ stimuli were generated. This constraint maximized the signal-to-noise ratio in the activity weights and improved the saliency of activity weights for a specific object category by only weighting clear positive and negative samples. 

Activity weights for training were generated using a modification of the $k$-fold cross validation technique. For a given training set that is a subset of the whole $1,386$ set, the collection of voxel data for the training set's images in the $646$-stimuli clear sample set was randomly split into five folds. For each of these folds, we held out the current fold as test data and combined the other four folds as training data. With this newly formed training set, a grid search was performed to tune the soft margin penalty parameter $C$ and RBF parameter $\gamma$ for an RBF SVM classifier using the LibSVM package~\cite{libsvm2011}. Finally, activity weights were generated by testing the classifier on the held-out fold to produce Platt probability scores~\cite{platt_2000} of class inclusion. This process was repeated for all five folds to generate activity weights for the collection of stimuli in the training set that are part of the clear sample set.

\subsection*{Experimental Design} Each of the original $500 \times 500$ colored images were down sampled to $250 \times 250$ grayscale images, with pixel values in the interval $[0,1]$. A layer of Gaussian noise with a mean of $0$ and variance of $0.01$ was added to each of these images. For each image, two feature descriptor types were independently generated. Histogram of Oriented Gradients (HOG) descriptors with a cell size of $32$ were generated using the VLFeat library's vl\_hog function~\cite{vedaldi08vlfeat}, which computes UoCTTI HOG features~\cite{felzenszwalb2010object}. Convolutional neural network (CNN) features were generated using the Caffe library's BLVC Reference CaffeNet model~\cite{DBLP:journals/corr/JiaSDKLGGD14}, which is AlexNet trained on ILSVRC 2012~\cite{russakovsky2014imagenet}, with minor differences from the version described by Krizhevsky et al.~\cite{krizhevsky2012imagenet}. Four partitions of training and test data were created. In each partition, $80\%$ of the data was randomly designated as training data and the remaining $20\%$ was designated as test data.

For each partition, experiments were conducted for the $127$ ways that the seven higher-level visual cortical regions (i.e., EBA, FFA, LO, OFA, PPA, RSC, and TOS) could be combined. In each experiment, for a given combination of higher-level visual cortical regions and for a given object category, two training steps were followed:

1. Activity weights were generated for a sampling of training stimuli, ones that are part of the 646-stimuli clear sample set, using an RBF-kernel SVM classifier trained on the training voxel data for that combination, following the fMRI activity weight calculation procedure described above.

2. Five balanced classification problems were created from the given partition's training data. For each balanced classification problem and each set of image descriptors (HOG and CNN features), two SVM classifiers were trained and tested --- one that uses a standard hinge loss (HL) function~\cite{Evgeniou_2000} and another that uses a activity weighted loss (AWL) function described by Equations~\ref{eqn:neuro_weighted_loss} \&~\ref{eqn:awl_mapping}. Both classifiers used an RBF-kernel.

The hinge loss function in Equation~\ref{eqn:hinge_loss} is solved via Sequential Minimal Optimization~\cite{export:68391}. It is not necessary to assign an activity weight $c_x \in C$ derived from fMRI data to every training sample; $c_x$ can be 0 to preserve the output of the original hinge loss function. In our experiments, $c_x \in [0,1]$, where $c_x$ corresponds to the probability that $x$ is in the object category in question; this results in penalizing more aggressively the misclassification of strong positive samples. The libSVM package was used to train and test SVM classifiers using a hinge loss function~\cite{libsvm2011}. To train classifiers using a activity weighted loss function, we modified publicly available code for an alternative additive loss formulation~\cite{Scheirer_2014_TPAMIa}. 

For each object category, combination of higher visual cortical regions, and set of image descriptors, we created five balanced classification problems. For each problem, we created a balanced training set with an equal number of positive and negative examples. For all object categories, because there were more negative than positive samples, all positive samples were used in the balanced problem and the same number of negative samples were randomly selected. The balanced problems only balanced the training data; each balanced problem used the same test set: the partition's held-out test set.

For both loss functions, binary SVM classifiers with RBF kernels were trained without any parameter tuning, using  parameters $C = 1$ and $\gamma$ = $(1 / \mathrm{number~of~features})$. The activity weighted loss function incorporates the calibrated probability scores from the first stage voxel classifiers as activity weights. We assigned these activity weights to the training samples that are members of the $646$-stimuli clear samples set. For samples without fMRI-derived activity weights, activity weights of $0.0$ are used. Finally, classifiers were tested on the partition's test set. In experiments using CNN features, RBF-kernel SVM classifiers converged during training, even though the vectors consisted of high-dimensional data.



\subsection*{Statistics for ROI Analysis} Because our analysis of the influence of specific ROIs involves comparing 127 quantities, to avoid multiple comparisons and to control for the family-wise error rate, Bonferroni correction was applied to adjust all confidence intervals. To create $m$ individual confidence intervals with a collective confidence interval of $1 - \alpha$, the adjusted confidence intervals were calculated calculated via $1 - (\alpha / m)$. With these adjusted confidence intervals ($m = 127$, $\alpha = 0.05$ and $\alpha = 0.01$), we compared the outputs of the empirical CDF function $F_X(x)$ for each null distribution $X$ that corresponded to an object category and set of image features as well as each ROI. 



\section*{Acknowledgements}
 The authors would like to thank J. Gallant and D. Stansbury for providing the fMRI data used in this study, as well as for helpful discussions regarding this work. This work was supported in part by NSF IIS Award \#1409097, IARPA contract \#D16PC00002, and an equipment donation from the NVIDIA Corporation. R. C. F. was supported in part by the Harvard College Research Program.
 
 \section*{Author Contributions Statement}
R.C.F., W.J.S., and D.D.C. designed research and wrote the manuscript text; R.C.F. performed research and prepared all figures.

\section*{Additional Information}
The authors declare no competing financial interests.

\bibliography{sample}

\end{document}